\documentclass{article}

\usepackage{microtype}
\usepackage{graphicx}
\usepackage{subcaption}
\usepackage{booktabs} 

\usepackage{hyperref}

\usepackage[preprint]{icml2026}

\usepackage{amsmath}
\usepackage{amssymb}
\usepackage{mathtools}
\usepackage{amsthm}

\usepackage[capitalize,noabbrev]{cleveref}

\theoremstyle{plain}

\theoremstyle{definition}

\theoremstyle{remark}

\usepackage[textsize=tiny]{todonotes}
\usepackage{algorithm,algorithmicx,algpseudocode}

\usepackage{multirow}
\icmltitlerunning{Style-CCL: Content-Preserving Style Transfer  via Curriculum Continual Learning}
\begin{document}

\twocolumn[
  \icmltitle{Style-CCL: Content-Preserving Style Transfer  via Curriculum Continual Learning}



  \icmlsetsymbol{equal}{*}

  \begin{icmlauthorlist}
    \icmlauthor{Shiwen Zhang}{teleai}
    \icmlauthor{Haoyuan Wang}{teleai}
    \icmlauthor{Xianghao Zang}{teleai}
    \icmlauthor{Haibin Huang}{teleai}
    \icmlauthor{Chi Zhang}{teleai}
    \icmlauthor{Xuelong Li}{teleai}
    
  \end{icmlauthorlist}

  \icmlaffiliation{teleai}{Institute of Artificial Intelligence (TeleAI), China Telecom}

  \icmlkeywords{Machine Learning, ICML}

  \vskip 0.3in
]



\printAffiliationsAndNotice{}  

\begin{figure*}[!htb]
  \centering
    \includegraphics[width=0.9\linewidth]{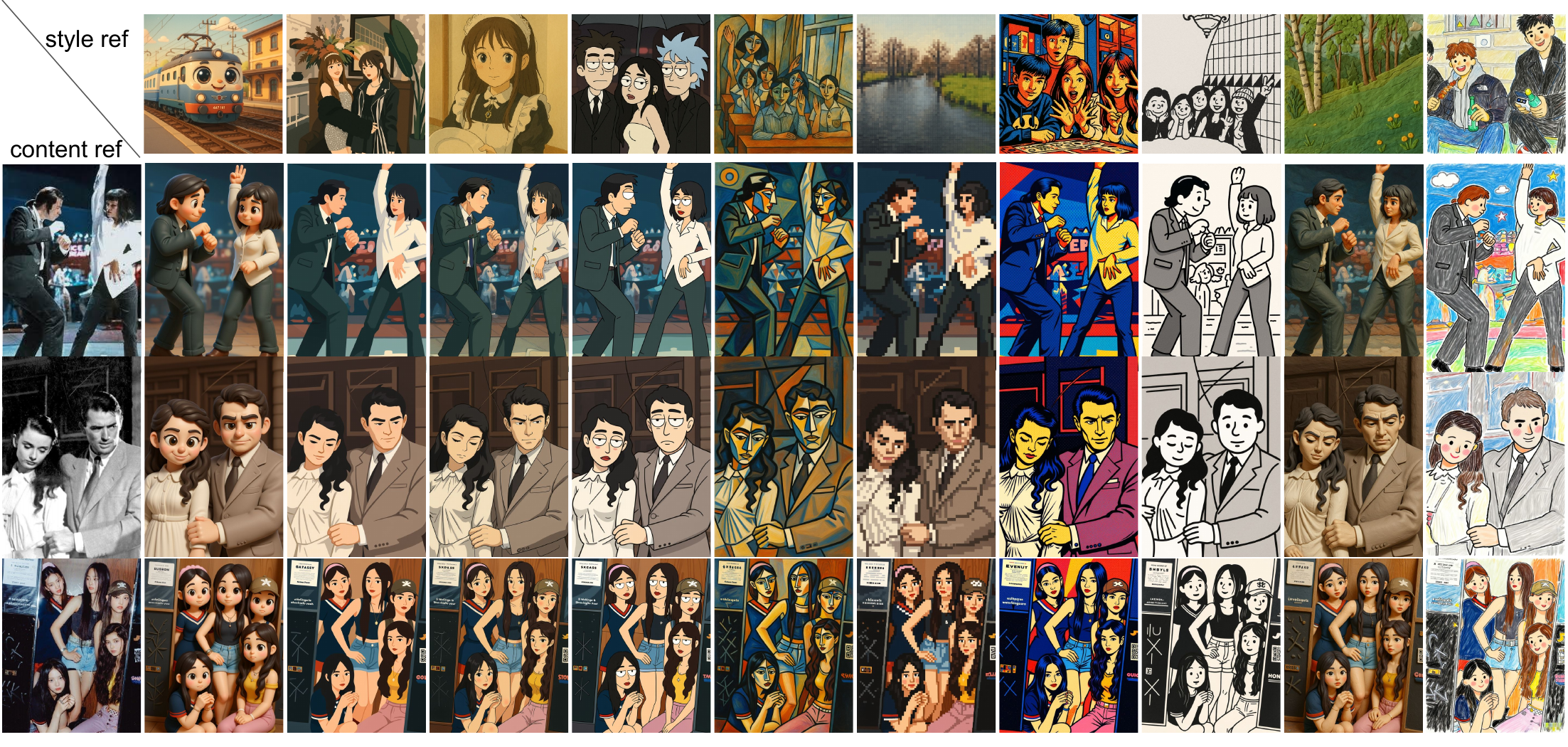}
    \captionsetup{type=figure}
    \captionof{figure}{ Style-CCL accepts style and content references for content-preserving style transfer, while maintaining high aesthetics merit.  }
    \label{figure_introduction}
\end{figure*}

\begin{abstract}

 Content-Preserving Style transfer, given content and style references, remains challenging for Diffusion Transformers (DiTs) due to entangled content and style features. With a reverse triplet synthesis pipeline to build a million-scale training set and a dual-branch Style-Content DiT (SC-DiT) that decouples style and content via separate ROPE embeddings and causal masking, we  observe that such a one-stage training paradigm on mixed style categories causes semantic styles to dominate, hindering texture style learning, and harming content preservation. To address these issues, we propose Style-CCL, a Multi-Stage Curriculum Continual Learning framework that trains SC-DiT from semantic (easy) to texture (hard) styles, and from clean to synthetic data, with Random Memory Rehearsal across stages to avoid catastrophic forgetting.  Extensive experiments  demonstrate that our Style-CCL achieves state-of-the-art performance in three core metrics: style similarity, content consistency, and aesthetic quality.
\end{abstract}
\section{Introduction}

 Image customization and editing with multiple references  \cite{wu2025qwen,flux-2-2025} with diffusion transformers \cite{flux2024,dit,esser2024scaling} has achieved great progress. However,  there is still significant scope for improving the effects of content-preserving style transfer \cite{gatys2016preserving}.
Current style transfer models \cite{wang2023styleadapter,zhang2025cdst} suffer from the leakage/invasion issue (subject/background/facial identities from style reference are over-transferred, polluting the characteristics of content reference) and struggle to keep multiple  characteristics in complex content reference. In addition, the generated images of style transfer models often exhibit low aesthetic merit.

In order to tackle aforementioned problems, we introduce decoupled style branch and content branch to DiT, termed SC-DiT ,  by utilizing VAE encoder \cite{DBLP:journals/corr/KingmaW13} to extract visual features for both branches and  apply  style RoPE and content RoPE \cite{rope}  to each branch to distinguish style and content with causal attention.  We collected and synthesized [style reference, content reference, target] triplets to train SC-DiT on FLUX-dev \cite{flux2024}.

However,  during training SC-DiT with various styles in one stage, we observe some surprising phenomena: First, semantics-related style transformations (e.g. 2D/3D cartoon, simple line drawings, vector design, etc) and texture-related style transformations (e.g. oilpainting, dense line drawings, texture materials, etc) contradict each other.  In particular, it turns out that semantics-related style transformations hinders the learning of  texture-related style transformations, even very long training time could not alleviate the issue.  Second, content characteristics fail to be well-preserved, as synthetic triplets compromise the integrity of clean triplets with respect to precise content preservation. 
Third,  for those particular style categories which contain both semantic-related and texture-related style transformations in the same style reference image, our model always learns semantic-related style transformation in early iterations and texture-related style transformations in late iterations. {\it Observation 3 clearly demonstrates that SC-DiT has the capability of learning texture-related styles when it is trained on only one style category, while Observation 1 indicates that such a capability is weakened and interfered when semantic and texture style categories are trained together. }

In order to tackle these problems, 
We propose a Multi-stage Style Curriculum Continual Learning (Style-CCL) paradigm. First, we introduce a theoratical tool, Local Intrinsic Dimensionality (LID) \cite{tempczyk2022lidl,kamkari2024geometric}, to estimate the complexity of style images. With a rough ranking of LID scores by our LID Estimator, we divide the style categories into semantic-related styles and texture-related styles with an approximate boundary. Then we apply our Style Curriculum Continual Learning to gradually learn these subsets from easy to hard, from clean to noisy \cite{bengio2009curriculum}, without catastrophic forgetting. We show the capability of our Style-CCL in Figure \ref{figure_introduction}. 

Our main contributions are:

\begin{enumerate}
    \item We observed that traditional one-stage training paradigm for  Conditional DiT with style and content references causes contradiction between semantic-related styles and texture-related styles, where semantic-related styles hinder the learning of texture-related styles. The characteristics of content reference is also not well-preserved in such one stage training.
    \item We propose a Multi-Stage Style Curriculum Continual Learning (Style-CCL) to tackle the aforementioned problems and  introduce Random Memory Rehearsal to avoid catastrophic forgetting. Our model could smoothly learn thousands of style categories with Style-CCL paradigm and preserve complex characteristics in content reference without subject confusion/mixture.
    \item Our Style-CCL achieves new state-of-the-art results in terms of style similarity, content preservation and aesthetics score through quantitative evaluation and user study.
\end{enumerate}

\section{Related Work}

\if
 Zero-Shot Style Transfer with  UNet Diffusion Models StyleID \cite{Chung_2024_CVPR} adapters Stable Diffusion \cite{Rombach2021HighResolutionIS} with training-free style injection. StyleShot \cite{gao2025styleshot} designs a style-aware encoder and a content-fusion encoder to conduct test-time-tuning-free style transfer. 
InstantStyle \cite{wang2024instantstyle} uses certain blocks of adapter in the inference process to avoid transferring content of style reference and requires an auxiliary ControlNet \cite{zhang2023adding} to support content reference. CSGO \cite{xing2024csgo} projects content reference into UNet Encoder and style reference into UNet Decoder with content-style image pairs for training.  
\fi

\paragraph{Zero-Shot Style Transfer with Conditional DiT}  SD3 \cite{esser2024scaling} and FLUX \cite{flux2024} improves text-to-image task significantly by scaling up DiT \cite{dit,dosovitskiy2020image,vaswani2017attention} parameters, outperforming previous UNet structures \cite{ddpm,Rombach2021HighResolutionIS,podell2023sdxl}. However, Such DiT models lacks disentangle properties \cite{zhangv4d, zhang2020knowledge,zhang2022tfcnet,zhang2023forgedit,zhang2025cdst,fastimagic,huang20204d,zhang2024hyper}. With these new powerful text-to-image DiT models, OminiControl \cite{tan2024ominicontrol} and EasyControl \cite{zhang2025easycontrol} enable Conditional Image Generation by concatenating condition image with text condition and noisy latent in self attention modules.   Qwen Image Edit \cite{wu2025qwen} could handle multiple reference images for subject-driven customization. However, by the time this paper was done (September, 2025),  Qwen Image Edit does not support subject+style references, neither does FLUX-Kontext \cite{labs2025flux}. OmniConsistency \cite{song2025omniconsistency} trained a separate content consistency branch and relies on external Style Loras \cite{lora} to conduct style transfer with content preservation. Instead, our model unifies content preservation and style transfer capability in one unified model, which is capable to handle universal style categories without the need for Style Loras trained on one specific style. 
\section{Style-CCL}
\subsection{Overview}
We begin by introducing our framework for constructing 
[style reference, content reference, target] training triplets. We then present the architecture of SC-DiT for content-preserving style transfer conditioned on both style and content references. Next, we describe three key empirical observations that expose fundamental limitations of the conventional one-stage training paradigm. Finally, we propose a curriculum continual learning strategy to address these issues. 
\subsection{Triplet Training Dataset Construction}

\begin{figure*}[!ht]
  \centering
  \hspace{-10mm}
   \includegraphics[width=0.8\linewidth]{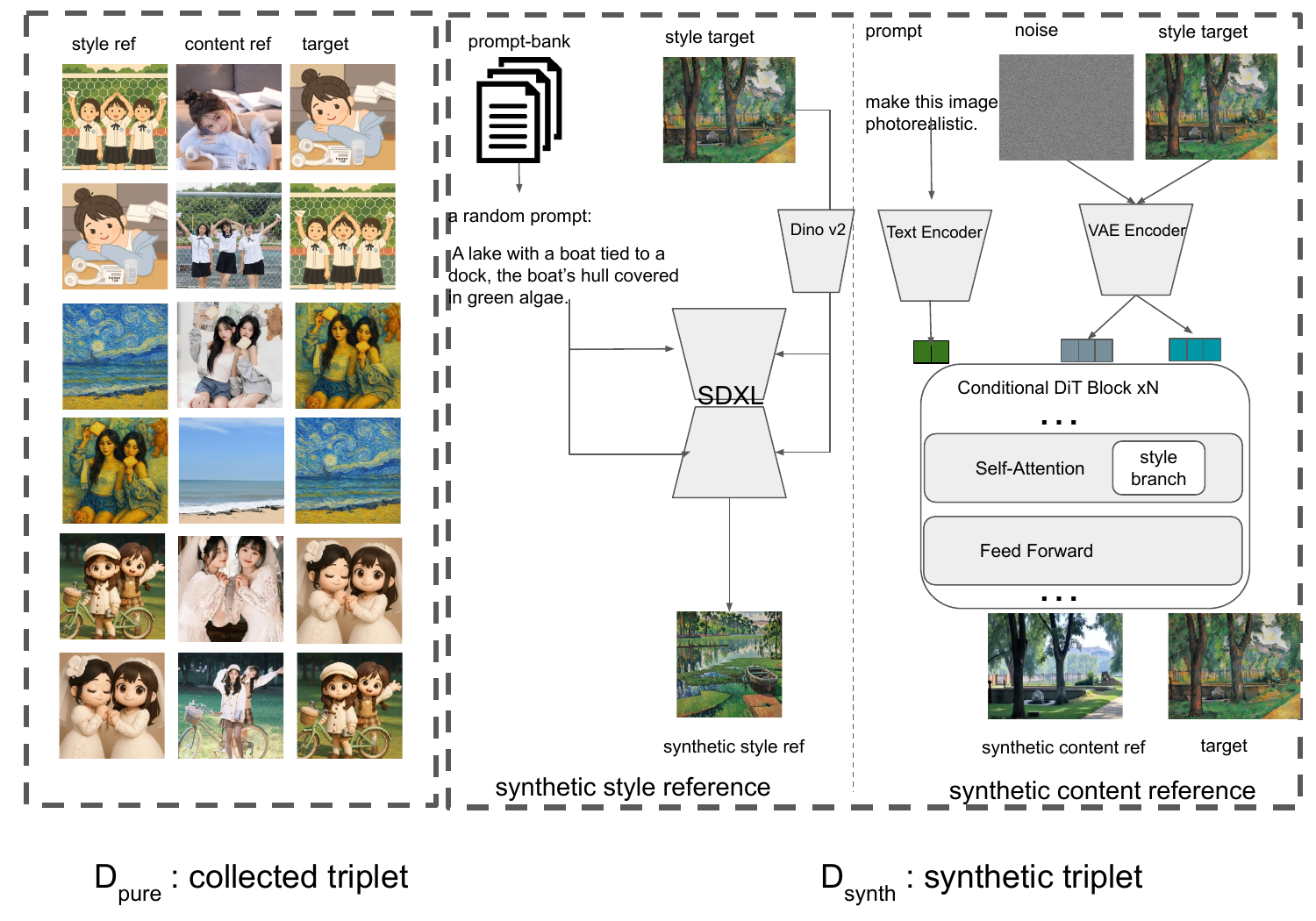}
   \caption{ Collected triplets $D_{pure}$ and  synthetic triplets $D_{synth}$  }
   \label{figure_triplet}
\end{figure*}

Unlike subject-driven image pair/triplet data, which naturally exists in videos or photo albums,  style triplets are rare in real world.
We collected [Style Ref, Content Ref, Target] image  triplets from a dataset  \cite{song2025omniconsistency} sampled from GPT-4O \cite{gpt4o} and some Loras from open-source community, and purified them with data cleaning. However, such collection is expensive and we only obtain 30 style categories. The model trained on such limited style categories generalizes poorly to unseen styles. Thus we introduce a reverse triplet synthetic framework inspired by \cite{wang2023stylediffusion} to generate training triplets from style images in-the-wild \cite{li2024styletokenizer}, where different style images are organized into noisy style clusters. The synthesis framework is shown in Figure \ref{figure_triplet}, where we specifically trained an image editing model on FLUX-dev to convert stylized image into photographic images.  Due to page limit, we elaborate implementation details  in the appendix. For simplicity, we call the collected clean triplet dataset $D_{pure}$ and the synthetic dataset $D_{synth}$ in the following sections. With a full matching strategy, we have around 330k triplets in $D_{pure}$ and 1 million triplets in $D_{synth}$, together containing more than 1k style  clusters.

\subsection{SC-DiT with Style Reference and Content Reference}

\begin{figure*}[t]
  \centering
   \includegraphics[width=1.0\linewidth]{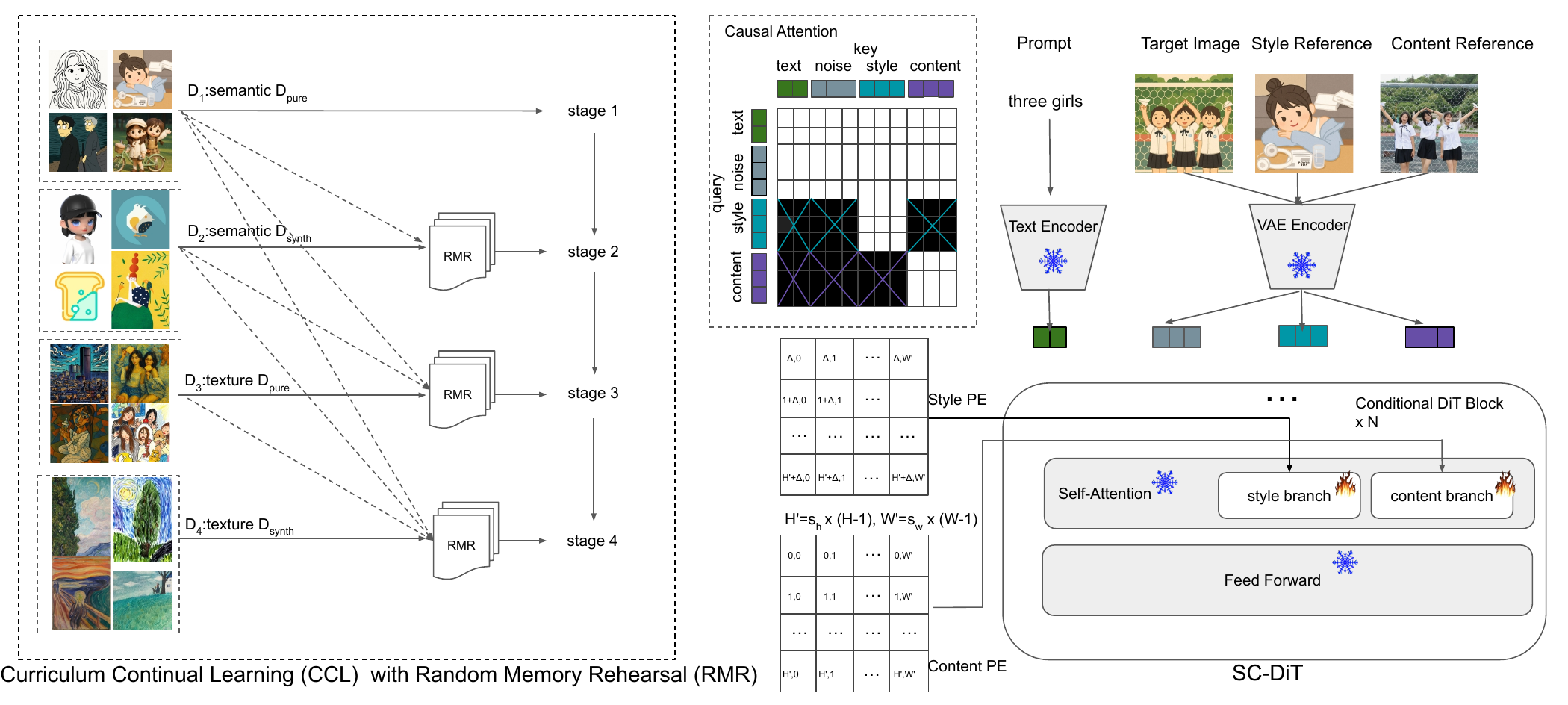}

   \caption{ Multi-Stage Style Curriculum Continual Learning and the structure of SC-DiT.  }
   \label{main}
\end{figure*}
We extend FLUX-dev \cite{flux2024} with a style-condition branch and a content-condition via separated Loras \cite{lora} and RoPE \cite{rope}, shown in Figure \ref{main}. Taking the query of  Double-Stream Block in FLUX for example, formally, we denote $Z_t, Z_n, Z_s, Z_c$ as intermediate representations of text, noise, style and content, respectively. $W_{Q_n}$ is the query matrix of the noisy image branch from the DoubleStreamBlock, with $W_{Q_t}$ being the text branch. We introduce $A_s,B_s$ as Lora for style injection, $A_c,B_c$ as Lora for content injection. They are attached to $W_{Q_n}$.

\begin{equation}
Q_t=W_{Q_t}Z_t, Q_n=W_{Q_n}Z_n
\label{attention}
\end{equation}
\begin{equation}
     Q_s=W_{Q_n}Z_s+B_sA_sZ_s, Q_c=W_{Q_n}Z_c+B_cA_cZ_c
\end{equation}
\begin{equation}
    {\bf Q}=[Q_t,Q_n,Q_s,Q_c]
\end{equation}
We have $Q_t,Q_n, Q_s, Q_c$ concatenated to form the overall query ${\bf Q}$. Although queries from each branch are concatenated,  Style Lora and Content Lora only operate on style feature and content feature themselves, without affecting  text branch and noise branch. Similarly, we could obtain the overall key ${\bf K}$ and value ${\bf V}$.  For SingleStreamBlock, the process is easier because there is only one $W_Q$ without distinguishing text and noise.

Then we apply Causal Attention with ${\bf Q,K,V}$ and a Causal Mask $M$, shown in Figure \ref{main}. With the mask $M$, we forbid the query from style and content  to text and noise, and forbid the interaction between style and content.  For $M$,  we set the white blocks  in Figure \ref{main} to 0,  black blocks in Figure \ref{main} to $-\infty$.

\begin{equation}
    O=softmax(\frac{{\bf Q}{\bf K}^T}{\sqrt{d}}+M){\bf V} 
\end{equation}

Furthermore, inspired by \cite{tan2024ominicontrol, zhang2025easycontrol}, we rescale the style reference of $H_s \times W_s$ and content reference of $H_c \times W_c$  to a fixed height $H$ and width $W$ with ratio:
\begin{equation}
    s^h_s=H_s/H,s^w_s=W_s/W,s^h_c=H_c/H,s^w_c=W_c/W
\end{equation}
we set the position encoding of content branch $PE_c$ and style branch $PE_s$ by:
\begin{equation}
    PE_s[i,j]=[s^h_s \times i+\Delta,s^w_s \times j]
\end{equation}
\begin{equation}
    PE_c[i,j]=[s^h_c \times i,s^w_c \times j]
\end{equation}
where we set $\Delta=H$ empirically. The  designs of $PE_s$ and $PE_c$ are different because the content reference should spatially align with the target yet the style reference should not, thus we add an offset to the style PE.

The optimization objective is based on rectified flow-matching \cite{rectifiedflow,flowmatching,esser2024scaling}:
\begin{equation}
L = \mathbb{E}_{t, \epsilon \sim \mathcal{N}(0, I)} \left\| v_\theta(x_t, t, c_{s}, c_{c}, c_{p}) - (\epsilon - x_0) \right\|_2^2
\end{equation}
where $x_t$ represents the image features at time $t$; $c_{s}$, $c_{c}$, and $c_{p}$ are the style, content, prompt conditioning inputs ; $v_\theta$ denotes the velocity field; $x_0$ is the target image feature; and $\epsilon$ is the noise.

\subsection{Three Observations }

\begin{figure*}[t]
  \centering
   \includegraphics[width=1\linewidth]{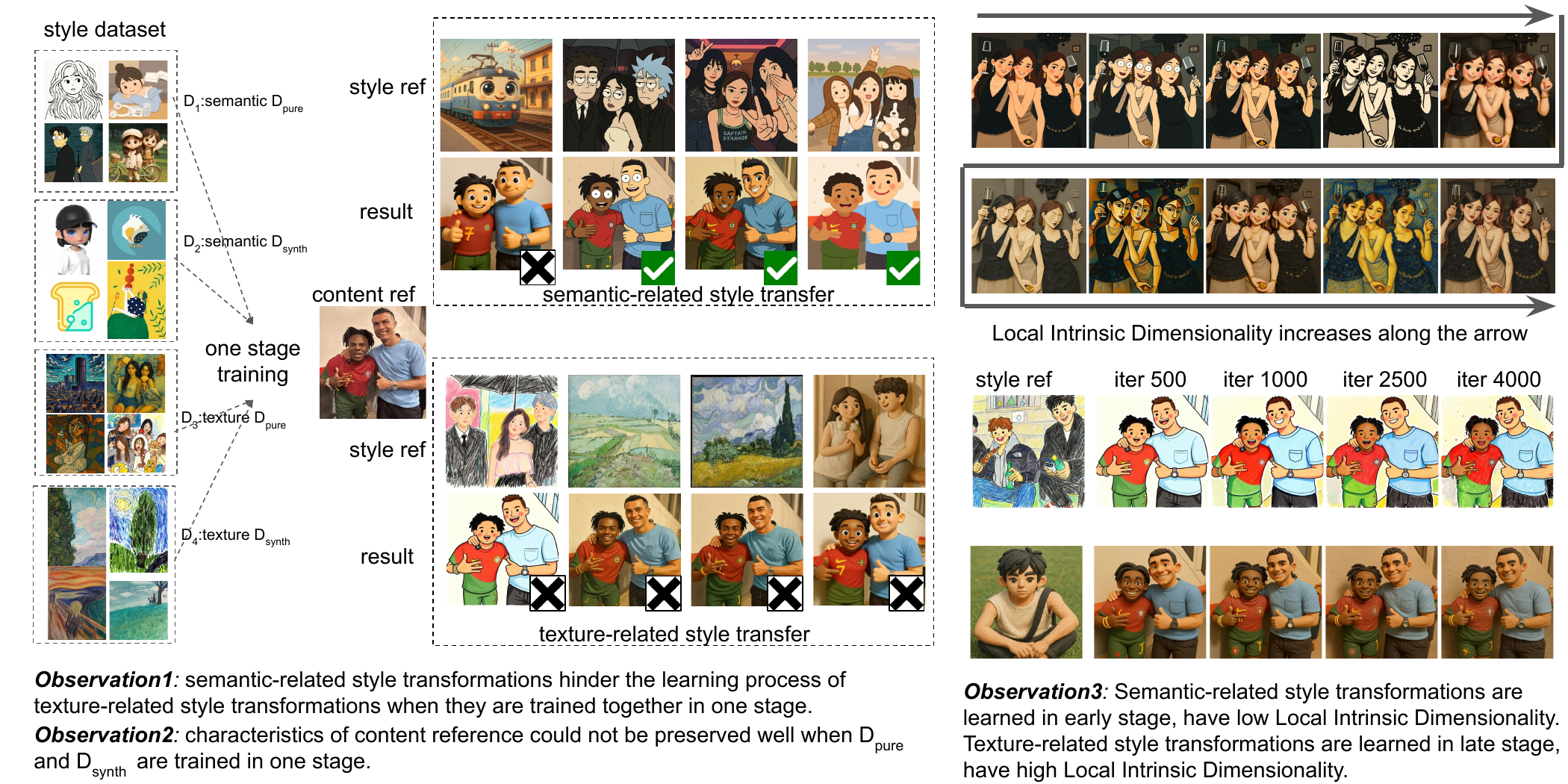}

   \caption{ Our key observations of SC-DiT.  }
   \label{figure_observe}
\end{figure*}

We initially trained SC-DiT in a single stage on a mixture of $D_{pure}$ and $D_{synth}$. However, we observed a surprising phenomenon: texture-related styles were consistently poorly learned, regardless of training time or data scale. For simplicity, we refer to styles without significant textures as semantic-related style transformations.  As illustrated in Observation 1 of Figure~\ref{figure_observe}, given the same content reference, semantic styles in the top row are transfered reasonably well, whereas texture styles in the bottom row are clearly deficient in strokes and textures (please zoom in for details). In the first column of the bottom row, prominent strokes are largely ignored. The second and third columns lack visible brushwork, and the fourth column appears overly smooth, missing the clay-like textures present in the style reference. We could also observed that the fine-grained characteristics could not be well-preserved in some cases, for example, the first column in the semantic-style row alters the skin color and clothes (the skew pattern is turned to horizontal) of the left person, the oilpainting in the texture-style could not preserve the facial identities of the content reference. 
Based on such evidence, we have\\
\textbf{ Observation 1:} 
{\it   Semantics-related style transformations hinder the learning of  texture-related style transformations when they are mixed and trained in one stage. } \\\textbf{ Observation 2:} 
{\it   Characteristics of content reference could not be preserved well when $D_{pure}$ and $D_{synth}$  are trained in one stage. }

We further quantitatively validate {\bf Observation 1} and {\bf Observation 2} in Table \ref{table_ablation_multistage}. To further investigate these phenomena,  we  selected styles that combine both semantic and texture transformations (right side of Figure~\ref{figure_observe}). In the first example, characters are transformed into a cartoon illustration (semantic) with strong pencil strokes (texture). In the second example, a 3D cartoon style (semantic) is combined with a clay material (texture). We trained SC-DiT on each of these styles \textit{individually} and inspected style transfer results at different iterations. We consistently found that semantic transformations are learned in early iterations, whereas texture transformations emerge in late stages. This experiment demonstrates that without significant interference from semantic styles, SC-DiT is capable of learning texture styles. It  also indicates that the complexity of  texture style is higher thus harder to learn than semantic styles.

\paragraph{Ranking Style Complexity with FPLID}Empirically, we choose to train an Fokker–Planck Local Intrinsic Dimensionality estimator (FPLID) ~\cite{kamkari2024geometric} in FLUX VAE space to approximately measure and rank the complexity of style images. We calculate the Spearman’s Correlation Coefficient and Significance on the sorted FPLID of a series of 20 style categories and 35 human users' averaged ranking results.  With  $\rho = 0.9718$, $p = 0.0007$, we confirm that LID ranking has a strong correlation with style complexity perceived by huamn.   Due to space limit, we elaborate the theoretical and experimental details of FPLID in the appendix . 
Concretely, we train a small DDPM U-Net~\cite{ddpm} on target images from $D_{pure}$ and $D_{synth}$ in FLUX VAE\cite{DBLP:journals/corr/KingmaW13} latent space:
\vspace{-0.2cm}
 \begin{equation}
 \resizebox{1\linewidth}{!}
 {
     $LID(x,t_0)=D-\sqrt{1 - {\bar{\alpha}_{t_0}}} tr (\nabla_x\epsilon(\sqrt{{\bar{\alpha}_{t_0}}}x, {t_0})) + \Vert {\epsilon}(\sqrt{{\bar{\alpha}_{t_0}}} x, {t_0})\Vert_2^2$
}
 \end{equation}

where $x$ denotes FLUX VAE Latent,  from $D_{pure}$ and $D_{synth}$, $t_0$ is the timestep for evaluating LID, with $D=16 \times 64 \times 64=65536$. With $\beta_t$ being the Diffusion process hyper-parameter, ${\alpha_t} := 1 - {\beta_t}$, ${\bar{\alpha}_t} := \prod_{s=1}^{{t}} {\alpha_t}$. The notation $tr$ denotes trace operation, $\nabla_x$ denotes  the differentiation operator with respect to $x$ and  $\epsilon$ is the Diffusion UNet.

We observe that semantic style transformations consistently exhibit lower LID, while texture style transformations have higher LID. Please zoom into the second row to see the fine textures on the skin in the last three images, especially the last one, where the style imitates canvas-like texture. Additional examples are provided in the appendix.Thus we obtain \\\textbf{Observation 3:} 
{\it   Semantic-related style transformations are learned in early stage, have low Local Intrinsic Dimensionality.  
Texture-related style transformations are learned in late stage, have high Local Intrinsic Dimensionality.  
 }

\subsection{Style Curriculum Continual Learning}
Inspired by our {\it Observation 3}, with collected dataset $D_{pure}$ and synthetic dataset $D_{synth}$, we  design a novel training paradigm, called {\bf Style Curriculum Continual Learning} (Style-CCL),  to tackle the severe problem of {\it Observation 1}, which causes our SC-DiT  performing poorly on texture-related styles.
{\it Observation 3} indicates that with VAE encoder to extract style features, semantic styles are easier to learn  and texture styles are harder to learn, which inspires us to apply curriculum learning to separate the training of semantic styles and texture styles. 

Shown in Figure \ref{main}, according to the sorted FPLID scores,  we divide $D_{pure}$ and $D_{synth}$ into four subsets, $D_1$: semantic styles from $D_{pure}$, $D_2$: semantic styles from $D_{synth}$, $D_3$: texture styles from $D_{pure}$, $D_4$: texture styles from $D_{synth}$. The division boundary of semantic and texture styles is manually set according to the sorted FPLID scores, The semantic/texture boundary of $D_{pure}$ and $D_{synth}$ are set separately. Due to the intrinsic vagueness of style definition, we do not  aim to and cannot precisely classify semantic and texture styles. Instead, the FPLID ranking and semantic/texture boundary are just helpful approximations to ease the training. We train $D_1$ and $D_2$ first, and train $D_3$ and $D_4$ later. However, the sequential multi-stage training on $\{D_1,D_2,D_3,D_4\}$ leads to catastrophic forgetting problem \cite{robins1995catastrophic}, with styles learned in early stages gradually forgotten in late stages. In addition, the content preservation is gradually weakened since the subset $D_1$ and $D_3$ from $D_{pure}$ have higher characteristics consistency than $D_2$ and $D_4$ from $D_{synth}$. Thus we introduce  Random Memory Rehearsal across Curriculum  Learning stages, by randomly sampling the same amount of  training data, with a fixed hyperparameter rehearsal rate $R$, from each style cluster in previous stages, and mix these previous samples with training data from current stage. The Random Memory Rehearsal is shown in Algorithm \ref{alg:rmr} and Style-CCL in Algorithm \ref{alg:Style-CCL}.

\begin{algorithm}[t]
  \caption{Random Memory Rehearsal (RMR)} \label{alg:rmr}
  \small
  \begin{algorithmic}[1]
    \vspace{.04in}
    \State{\bf Input}: 
    The previous triplet dataset $D_{pre}=\{S_1,S_2,...,S_{N_{pre}}\}$ containing $N_{pre}$ style clusters,  and the current triplet dataset $D_{cur}=\{S'_1,S'_2,...,S'_{N_{cur}}\}$ containing $N_{cur}$ style clusters, and  a fixed rehearsal sampling rate $R$.

    {\bf Output}:  $D_{rmr}=\{T_1,T_2,...,T_{N_{pre}},S'_1, S'_2,...,S'_{N_{cur}}\}$ containing $N_{rmr}=N_{pre}+N_{cur}$ style clusters.

    \State {\bf Procedure}:

    \State $D_{rmr} \leftarrow \{\}$ 
    \For{$S_i \in D_{pre}$}
      \State Given a fixed rehearsal sampling rate $R$, we have sampling number $k$ for each style cluster,  $k \leftarrow \sum\limits_i^{N_{cur}}|S'_i| \times R/ N_{pre}$ 
      \State Randomly sample triplets $T_i=\{s_i^1,s_i^2,...,s_i^k\}$ from $S_i$
      \State Insert $T_i$ to $D_{rmr}$
    \EndFor
    \State \textbf{return} $D_{rmr}$
    \vspace{.04in}
  \end{algorithmic}
\end{algorithm}


\begin{algorithm}[t]
  \caption{Style Curriculum Continual Learning (Style-CCL)} \label{alg:Style-CCL}
  \small
  \begin{algorithmic}[1]
    \vspace{.04in}
    \State {\bf Input}:\\ $D_1$: semantic styles from $D_{pure}$, \\$D_2$: semantic styles from $D_{synth}$, \\$D_3$: texture styles from $D_{pure}$, \\
    $D_4$: texture styles from $D_{synth}$.
    \State {\bf Output}:  SC-DiT$_{final}$
    \State
    
    \State {\bf Procedure}:
    \State Train SC-DiT$_1 \leftarrow \{D_1,$ FLUX-dev$\}$
    \State Train SC-DiT$_2 \leftarrow \{$RMR($D_1,D_2$), SC-DiT$_1\}$
    \State Train SC-DiT$_3 \leftarrow \{$RMR($D_1,D_2,D_3$), SC-DiT$_2\}$
    \State Train SC-DiT$_4 \leftarrow \{$RMR($D_1,D_2,D_3,D_4$), SC-DiT$_3\}$
    \State SC-DiT$_{final} \leftarrow$ SC-DIT$_4$
    \State \textbf{return}  SC-DiT$_{final}$
    
  \end{algorithmic}
 
\end{algorithm}

\section{Experiments}

\paragraph{Implementation Details.} We adopt FLUX dev 1.0 \cite{flux2024} as base model for SC-DiT. The ranks for Style Branch Lora and Content Branch Lora are 128. We apply gradient checkpointing \cite{griewank2000algorithm} to save memory thus our model could easily be trained with short sides of both $512 \times$ and $1024 \times $. Our model is trained with 4 H100 GPUs, batch size is 1 for each GPU, learning rate is 1e-4.

\paragraph{Evaluation Benchmark.}
We select 50 style references and 40 content references, mutually pair each of them to generate 2000 style-content pairs for testing.  We further select 10 style references and 10 content references as  validation set.  The style references cover diverse  style genres and the content references include different number of persons with diverse gestures, scenes/buildings and subjects in complex scenarios. The images are of different aspect ratios. we show the details of these benchmarks in the appendix. 
\paragraph{Evaluation Metrics.} 
We evaluate our method with the following metrics. 
For \textbf{Style Consistency}, we use  CSD Score~\cite{csd} to measure the style similarity between the style reference and the generated image. 
For \textbf{Aesthetics},we use the LAION Aesthetics Predictor~\cite{schuhmann2022laion} to estimate the aesthetic quality of the generated image. For \textbf{Content Preservation},  we propose a new \emph{Content Preservation Cut-Off Score} (CPC Score) with a style consistency threshold. Intuitively, a model that simply replicates the content reference without transferring style would receive an artificially high content score. To avoid this, we first use Qwen-VL~\cite{bai2025qwen2} to generate a detailed caption $T_\text{vlm}$ for the content reference image $I_{\text{content}}$, and compute the CLIP score~\cite{Radford2021LearningTV} between $T_{\text{vlm}}$ and the generated image $I_{\text{res}}$. We then compute the CSD Score between $I_{\text{style}}$ and $I_{\text{res}}$; if this score falls below a threshold, the CLIP score is set to zero as a penalty. 

\begin{equation}
\resizebox{0.9\columnwidth}{!}{%
$CPC@thresh =
\begin{cases}
CLIP(I_{res},T_{vlm}), & \text{if }  CSD(I_{res},I_{style})>=\text{thresh} \\
0, & \text{if } CSD(I_{res},I_{style})<\text{thresh} \\

\end{cases}$
}
\end{equation}

\begin{figure*}[!ht]
  \centering
   \includegraphics[width=0.8\linewidth]{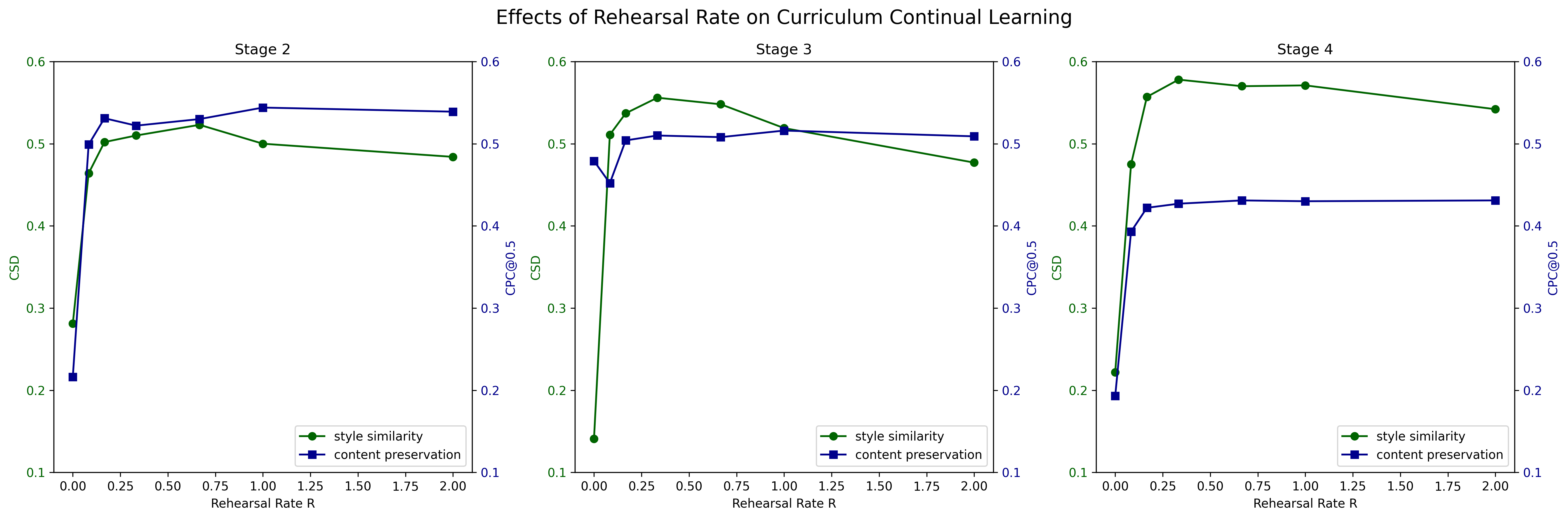}
   \caption{ Quantitative ablation studies on the effects of Continual Learning and Rehearsal Rate. We train 7 models with different Rehearsal Rates $R$ and validate the CSD Score and CPC Score on the validation set of each stage. When $R=0$, no Curriculum Learning is applied.   }
   \label{figure_rehearsal}
\end{figure*}

\subsection{Ablation Study}

\begin{figure}[!ht]
  \centering
   \includegraphics[width=0.8\linewidth]{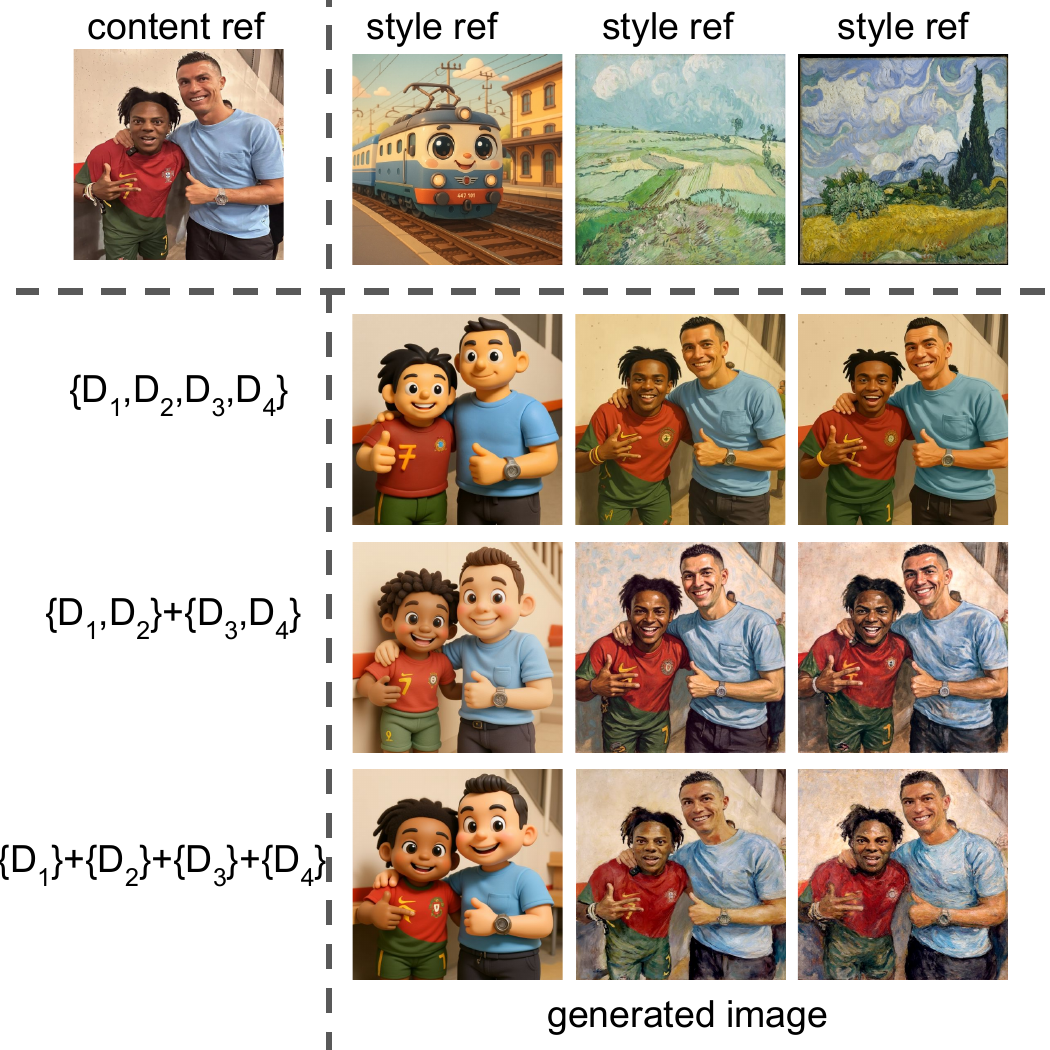}

   \caption{Importance of Multi-Stage CCL. }
   \label{figure_ablation_multistageccl}
\end{figure}

\paragraph{Importance of Multi-Stage CCL}
We conduct quantitative evaluations for Observation 1 and Observation 2 in Table \ref{table_ablation_multistage}, demonstrating that texture styles interfere the semantic styles. Shown in Figure \ref{figure_ablation_multistageccl}, we qualitatively compare the effects of number of CCL stages.
\begin{table}[!htb]

\begin{center}
\resizebox{1.05\columnwidth}{!}{%

\begin{tabular}{c|c|c|c|c}
\hline
{Training Strategy} &  {Semantic Style $\uparrow$} & {Texture Style $\uparrow$} &  {Overall Style  $\uparrow$}  & {Content Preservation  $\uparrow$}  \\
\hline
One Stage &0.571 & 0.117 & 0.344 & 0.298 \\
\hline
Two Stages & 0.574 & 0.526 &0.557 & 0.392 \\
\hline
Four Stages &{\bf 0.595} &{\bf 0.561} & {\bf 0.578}& {\bf 0.427}\\

\hline
\end{tabular}
}
\end{center}

\caption{Quantitative ablation studies on  the multi-stage CCL training strategy. We found that one stage training with mixed semantic and texture styles causes low style similarity for texture styles, which quantitatively validates our Observation 1.}
\label{table_ablation_multistage}
\end{table}

When SC-DiT is trained in one stage with a naive mixture of $\{D_1,D_2,D_3,D_4\}$, we could find in Figure \ref{figure_ablation_multistageccl}, the T-shirt of the left person is inconsistent with the content reference (the green part of the T-shirt should be skew) and his skin color is not correctly preserved. The oil painting styles could not be correctly illustrated   and characteristics of these two persons  obviously change.

When SC-DiT is trained in two stage CCL $\{D_1,D_2\}+\{D_3,D_4\}$ (we apply Random Memory Rehearsal in the second stage, which is not explicitly written in Figure \ref{figure_ablation_multistageccl} for simplicity), we could find that SC-DiT could simultaneously learn semantic styles and texture styles. However, the characteristics still could not be preserved well.  The green part of the T-shirt, worn by the left 3D-cartoon person, is still not skew. The facial identities of the persons in oil painting results are still not similar enough with the content reference. This is due to fact that clean collected triplets and noisy synthetic triplets are trained together, which has a negative impact on the characteristics consistency.
\begin{figure*}[t]
  \centering
   \includegraphics[width=0.9\linewidth]{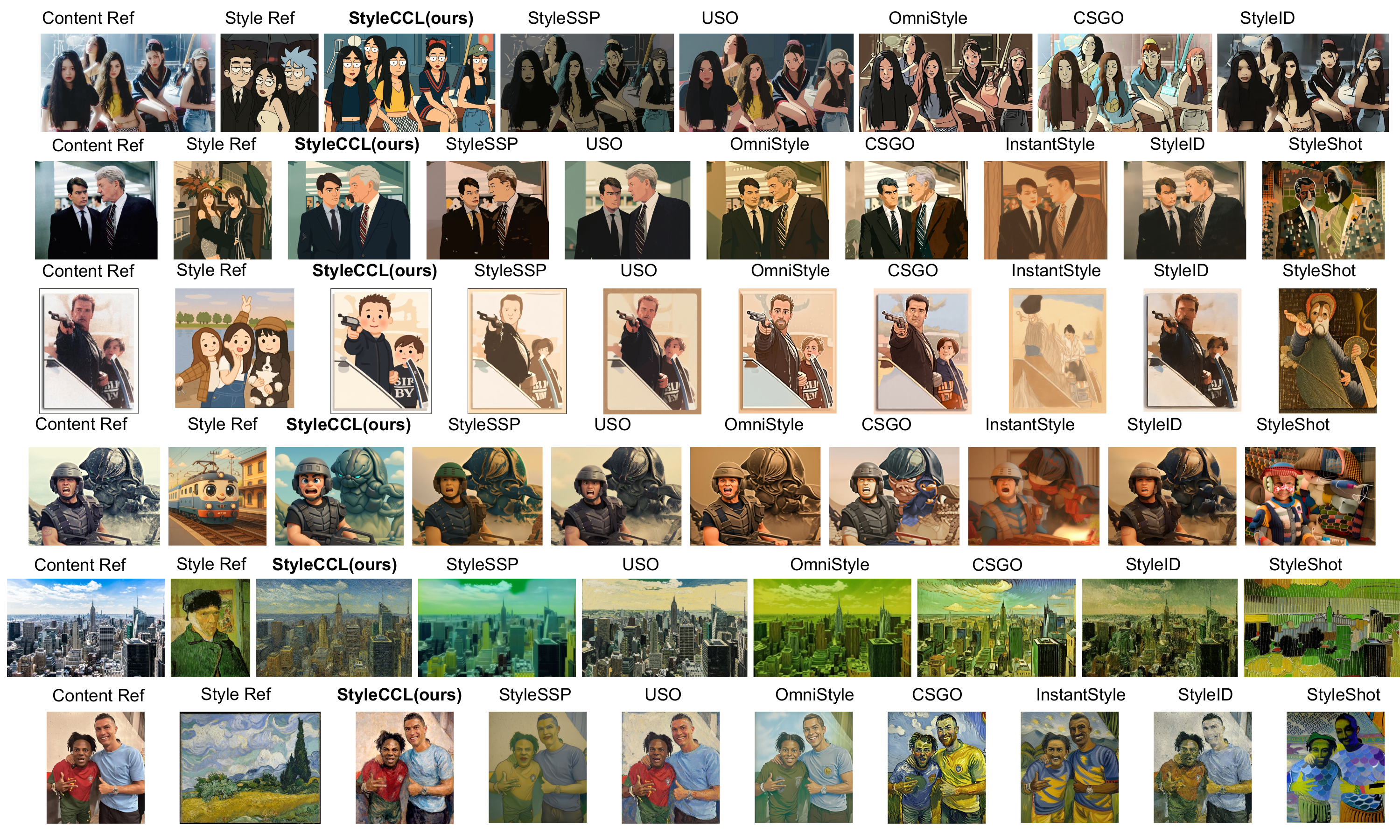}
   \vspace{-0.4cm}

   \caption{ Qualitative Comparison with State-of-the-art Style Transfer Models.   }
   \label{figure_sota}
\end{figure*}

When SC-DiT is trained with four-stage CCL $\{D_1\}+\{D_2\}+\{D_3\}+\{D_4\}$ ( algorithm \ref{alg:Style-CCL}), we could find SC-DiT performs well on both semantic styles and texture styles. Furthermore, the characteristics of content references could be well-preserved.

\begin{table*}[!htb]
\vspace{-0.3cm}
\scriptsize
\begin{center}
\begin{tabular}{c|c|c|c|c}
\hline
{Model} &  {Style Similarity CSD Score$\uparrow$}  & {Content Preservation CPC Score@0.5 $\uparrow$}  & {Content Preservation CPC Score@0.3:0.9 $\uparrow$}  & {Aesthetic Score$\uparrow$}  \\
\hline

OmniStyle& {0.447} & 0.194 & 0.163 & 5.881  \\
OmniGen-v2 & 0.462 & 0.243 & 0.166& 5.843\\
DreamO & 0.402 & 0.193 & 0.102 & 6.149 \\
StyleID & 0.453 & 0.190 & 0.180& 5.749 \\
StyleShot & 0.450 & 0.227 & 0.116& 5.740 \\
InstantStyle  & 0.397 & 0.189 &0.134& 5.464 \\
StyleSSP & 0.494 & 0.291 & 0.207&5.130 \\
CSGO  & \underline{0.535} & \underline{0.379} & \underline{0.224} & 5.969 \\

\hline

{\bf Style-CCL (ours)} & {\bf 0.561} & {\bf 0.401} & {\bf 0.236}&  {\bf 6.297} \\

\hline
\end{tabular}
\end{center}
\caption{Quantitative comparison of our Style-CCL with previous state-of-the-art style transfer methods. The best score is stressed by bold font and the second best score is marked by underline.}
\vspace{-0.5cm}
\label{table_sota}
\end{table*}

\textbf{Continual Learning and Random Memory Rehearsal} We thoroughly explore the effects of Rehearsal Rate $R$ of Style Curriculum Continual Learning. We quantitatively measure style similarity with CSD Score and content preservation with CPC Score on the validation set of each stage in Table \ref{table_ablation_rehearsal_rate}. We found that with an increasing $R$, the style similarity first increases and then decreases, while the content preservation keeps increasing then gradually saturates. We choose to set the Rehearsal Rate $R$ to $1/3$.

We plot the effects of Random Rehearsal Rate $R$ of each stage in Style-CCL  on the validation set in Figure \ref{figure_rehearsal}. We do not show the  first stage because it is a normal training without continual learning. We observe some interesting phenomena in Figure \ref{figure_rehearsal}:
\begin{itemize}
    \item  With an increasing $R$, the style similarity first increases then decreases. When $R=0$ there is no continual learning thus some styles in previous stage will be forgotten, which leads to very low style similarity in every stage. However, when $R$ keeps increasing, style data from previous stages dominate, which hinders the learning of data from current stage, thus style similarity gradually decreases.
    \item With an increasing $R$, the trending of content preservation is different for different stages. In stage 2 and stage 4, where noisy synthetic data dominates the current stage, we could observe content preservation keeps increasing with $R$ and gradually saturates. In stage 3, where clean data dominates the current stage, increasing $R$ leads to no significant fluctuation. 

\end{itemize}

\begin{figure*}[!htp]
  \centering
  \hspace{-10mm}
   \includegraphics[width=0.77\linewidth]{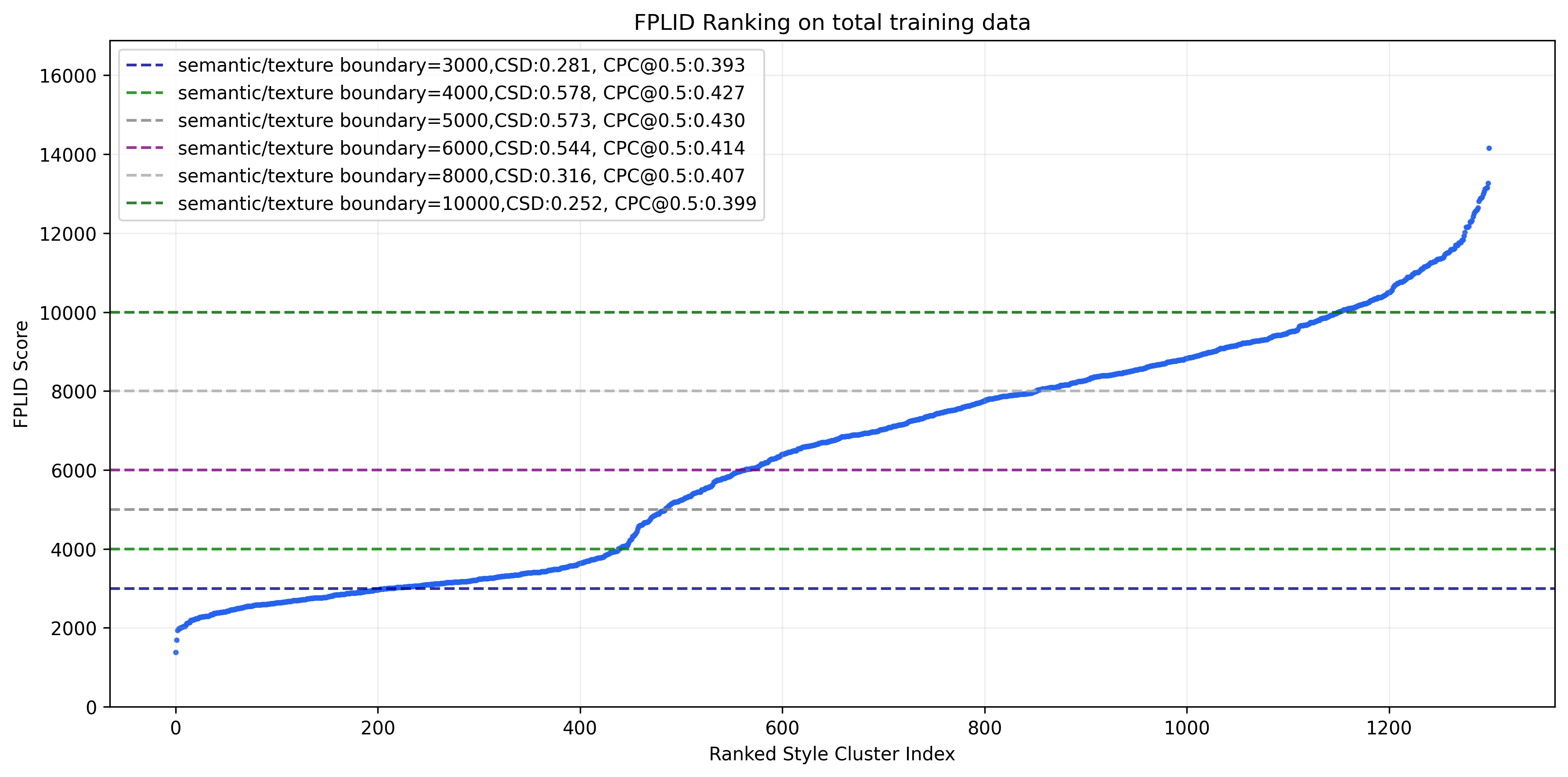}
   \caption{ We rank all style clusters by the average of FPLID in each cluster. Various semantic/texture boundaries are experimented, and we epirically set it to be 4000. }
   \label{figure_stboundary}
\end{figure*}

\textbf{Semantic Texture Boundary} We rank the FPLID score of all style clusters in training set by averaging each cluster. We train a model with semantic/texture data with style-ccl for each boundary and measure the models' performance with CSD score and CPC@0.5 score on validation set. Shown in Figure \ref{figure_stboundary}, we epirically set the semantic-texture boundary to FPLID=4000. This is never meant to be a precise division. Instead, it is just to ease the training process for Style-CCL.

\subsection{Comparison with State-of-the-art Methods}

\paragraph{Quantitative Comparison}
We quantitatively compare our Style-CCL with multiple current state-of-the-art style transfer models in Table \ref{table_sota}, including UNet-based, DiT-based from the aspects of style similarity, content preservation and aesthetics score.\\
\textbf{Qualitative Comparison}
We present qualitative visual comparison with state-of-the-art style transfer models  on diverse style references from our test benchmark in Figure \ref{figure_sota}, where our Style-CCL tackles both semantic-related and texture-related style transfer and generate images with high aesthetics. 

\begin{table}[t]
\begin{center}
\resizebox{0.9\columnwidth}{!}{%
\begin{tabular}{c|c|c|c|c}
\hline
{Model} &  {Style}  & {Content }  & {Aesthetics} & {Overall} \\
\hline

StyleShot & 0.75\% & 0.25\% & 1.75\% & 0.25\%\\
InstantStyle  & 3.25\% & 1.25\% & 0.25\% &  0.50\%\\

StyleSSP & 3.75\% & 4.50\% & 8.25\% & 2.25\%\\
StyleID & 2.00\% & 22.25\% & 2.50\% & 3.75\% \\
CSGO  & 5.50\% & 6.25\% & 1.00\% & 5.25\%\\
OmniStyle& {6.50\%} & 7.50\% & 4.25\% & 5.50\%  \\

USO & 8.50\% & 27.75\% &10.00\% & 9.75\%\\

\hline

{\bf Style-CCL (ours)} & {\bf 69.75\%} & {\bf 30.25\%} &  {\bf 72.00\%} & {\bf 72.75\%}\\

\hline
\end{tabular}
}
\end{center}
\vspace{-0.1cm}
\caption{User Study.}
\vspace{-4mm}
\label{table_userstudy}

\end{table}

\textbf{User Study.}
We employ 20 human evaluators to pick one best performance model from the candidates regarding style similarity, content consistency, aesthetics and their  overall choices. The user study result is presented in Table \ref{table_userstudy} in percentage format. 

\subsection{Limitations of Style-CCL-FLUX 1.0} 
Please note these limitations are for Style-CCL FLUX only. We observe two  main limitations, as shown in Figure~\ref{figure_limations}. 
First, in crowded scenes with many people, it may fail to preserve all individuals or their characteristics and it also struggles with rare, fine-grained out-of-distribution styles (e.g., Chinese ceramic art).  In fact, our QwenStyle \cite{qwenstyle} and TeleStyle series \cite {telestyle, telestylev2} have tackled these issues.

\section{Transfer Style-CCL to stronger foundation models}
\begin{figure*}[t]
  \centering
   \includegraphics[width=1\linewidth]{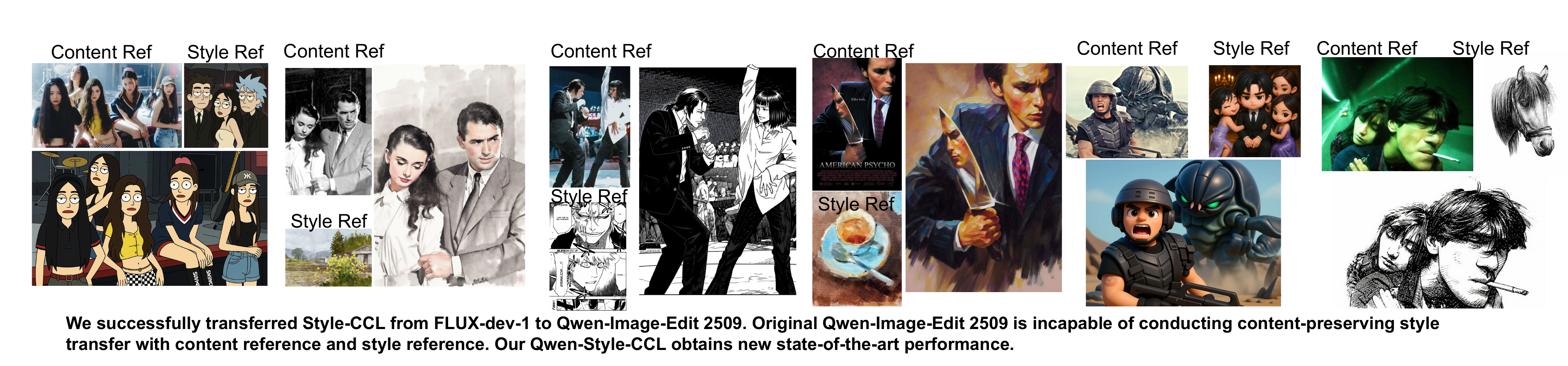}

   \caption{ We transfer Style-CCL algorithm to Qwen-Image-Edit-2509 to train QwenStyle. The vanilla QIE-2509 is incapable of content-preserving style transfer. Our QwenStyle established new state-of-the-art on this task. }
   \label{figure_qwenstyle}
\end{figure*}

\begin{figure*}[t]
  \centering
   \includegraphics[width=0.8\linewidth]{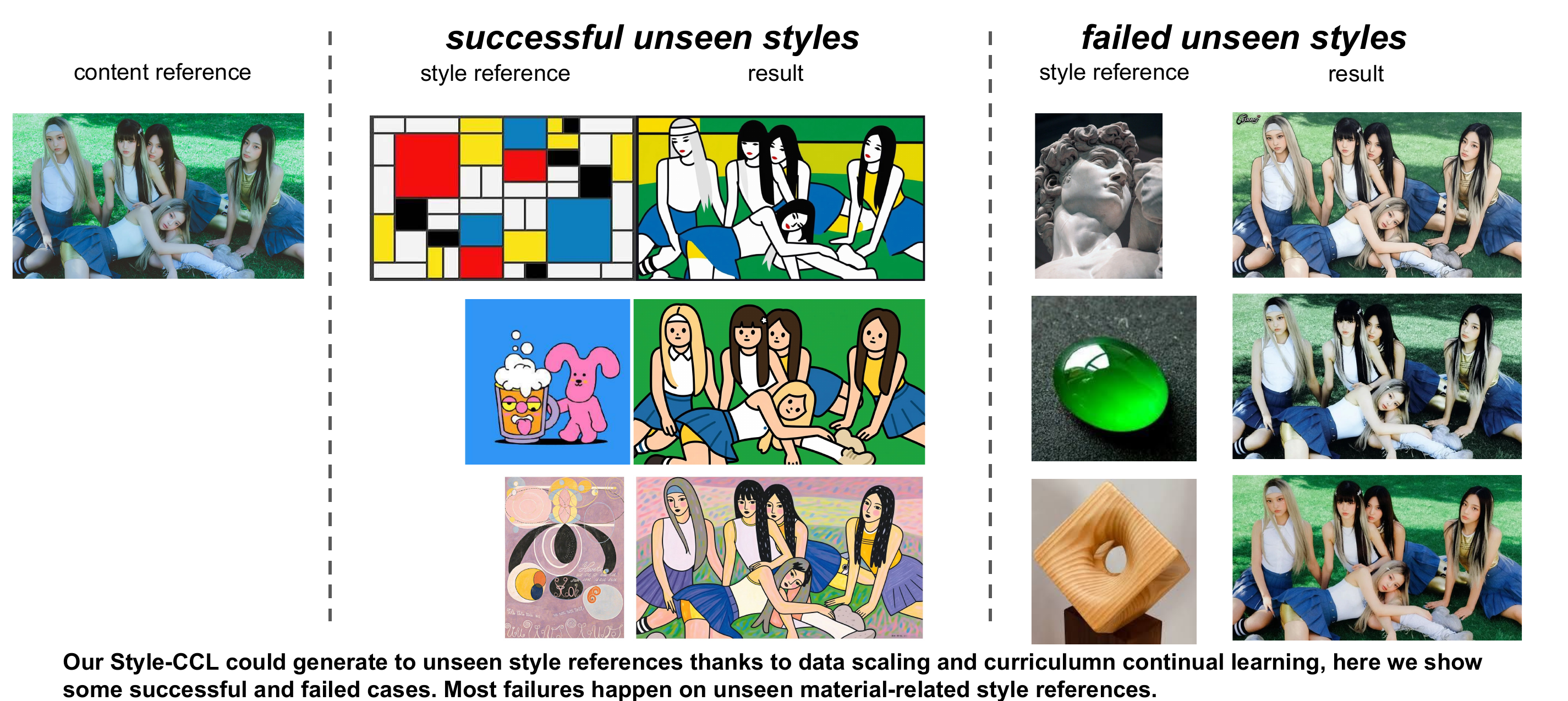}

   \caption{ Out-of-Distribution cases for QwenStyle. The "failed" cases are not even wrong, since these style references could be interpreted as "photo-realism" styles. If one needs to transfer the material, perhaps it is better to use prompt directly.}
   \label{figure_ood}
\end{figure*}

We have successfully transfer Style-CCL to Qwen-Image-Edit series (2509, 2511) \cite{wu2025qwen} in December 2025. Our first version of Style-CCL model is TeleStyle V1 (QwenStyle) \cite{qwenstyle,telestyle}, released and open-sourced in Jan 2026, demonstrating strong generalization capability, high style similarity, content consistency and aesthetic merits. TeleStyle V1 (QwenStyle) established new state-of-the-art content-preserving style transfer performance in open-source models. We show a few examples in Figure \ref{figure_qwenstyle}. QwenStyle could generalize to unseen styles in Figure \ref{figure_ood}. In June 2026, we release TeleStyle V2 \cite{telestylev2}, achieving style transfer performance on par with top close-source model, gemini-3-pro-image-preview (nano banana pro) \cite{gemini3pro}. Beyond content-preserving style transfer task, TeleStyle series are also general text-guided image editing models on par with Qwen-Image-Edit \cite{wu2025qwen} via Distribution-Matching-Distillation \cite{dmd2,phasedmd}.

\section{Conclusion}
We observed that the one-stage training paradigm of SC-DiT suffers from semantic-texture interference and characteristics shifting. Thus we present Style-CCL, a Multi-Stage Style Curriculum Continual Learning framework content-preserving style transfer to tackle these problems.  Our Style-CCL achieves new state-of-the-art performance on style similarity, content preservation and aesthetics score.

\section{Appendix}

\subsection{Limitations of Style-CCL-FLUX 1.0}
\begin{figure}[htb]
  \centering
  \hspace{-10mm}
   \includegraphics[width=1\linewidth]{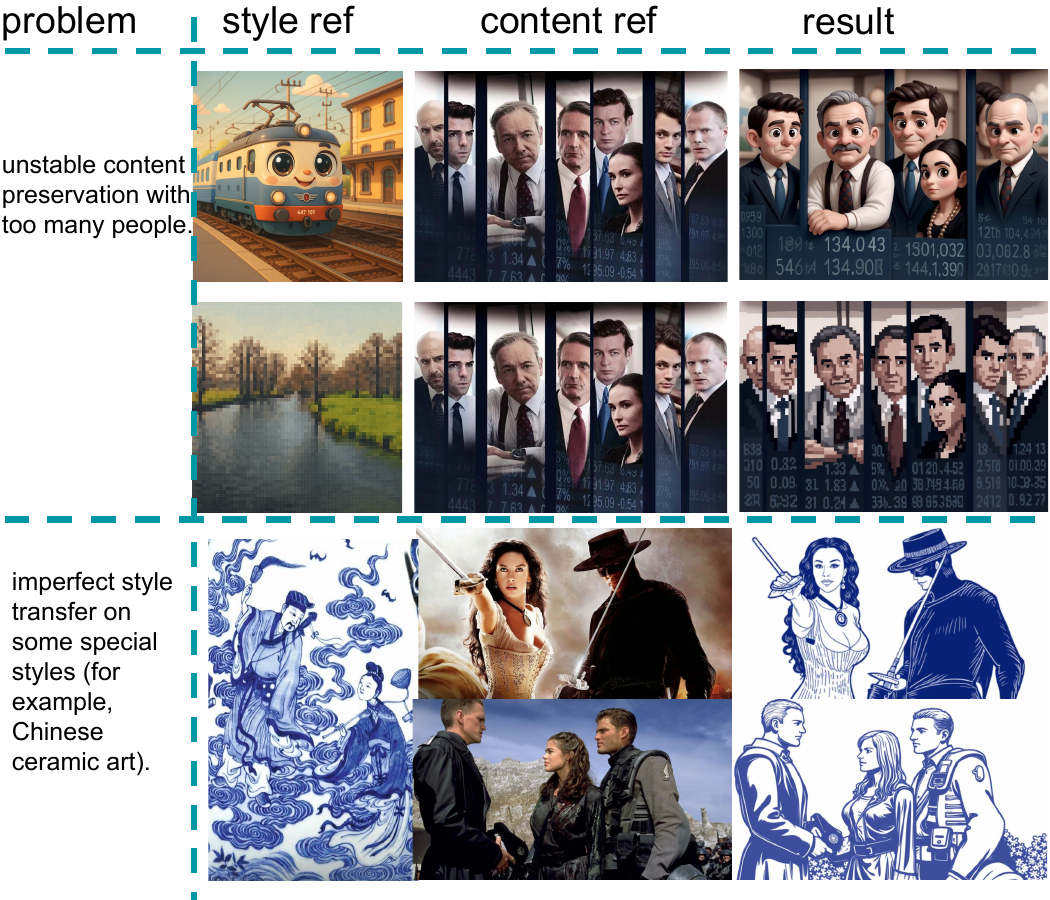}

   \caption{Limations of Style-CCL-Flux 1.0: Our model is unstable when there are too many people in the content reference and style fidelity decreases on some specific style genres.  }
   \label{figure_limations}
\end{figure}

We observe two  main limitations, as shown in Figure~\ref{figure_limations}. 
First, in crowded scenes with many people, our model could be unstable thus may fail to preserve all individuals or their characteristics from the content reference. For example, in the first row, the 3D cartoon style  could not preserve the correct number of characters. However, in the second row, the pixel effect could maintain the correct number of people. Second, our model may also struggle with some fine-grained out-of-distribution styles. For example, the third and fourth rows demonstrate the style transfer effects with Chinese ceramic art style reference. The style could not be precisely reproduced and looks more like a comic-book style,though the  color is corretly transferred.

\subsection{Quantitative experiments on Multi-Stage CCL and Rehearsal Ratio}
We show the quantitative ablation of rehearsal rate in Table \ref{table_ablation_rehearsal_rate}

\begin{table*}[!htb]
\begin{center}
\resizebox{2\columnwidth}{!}
{
\begin{tabular}{c|c|c|c|c|c|c|c|c}
\hline

&{Rehearsal Ratio $R$ }  & {0} & {1/12} & 1/6 & 1/3 & 2/3 & 1 & 2\\\cline{1-9}
\multirow{2}{*}{Stage 2}
&{Style Similarity on $D_1+D_2 \uparrow$ }& 0.281 & 0.464 & 0.502 & 0.510 & \textcolor{teal}{\bf  0.523} & 0.500 & 0.484\\
\cline{2-9}
&{Content Preservation on $D_1+D_2 \uparrow$}& 0.216 &0.499 & 0.531 &0.522&0.530 & \textcolor{teal}{\bf 0.544} &0.539 \\
\cline{2-9}

\hline
\multirow{2}{*}{Stage 3}
&{Style Similarity on $D_1+D_2+D_3 \uparrow$}& 0.141 & 0.511 & 0.537 & \textcolor{teal}{\bf 0.556} & 0.548 & 0.519 & 0.477\\
\cline{2-9}
&{Content Preservation on $D_1+D_2+D_3 \uparrow$}& 0.479 &0.452 & 0.504 &0.510&0.508 & \textcolor{teal}{\bf 0.516} &0.509 \\
\cline{2-9}

\hline
\multirow{2}{*}{Stage 4}
&{Style Similarity on $D_1+D_2+D_3+D_4 \uparrow$}& 0.222 & 0.475 & 0.557 & \textcolor{teal}{\bf 0.578} & 0.570 & 0.571 & 0.542\\
\cline{2-9}
&{Content Preservation on $D_1+D_2+D_3+D_4 \uparrow$}& 0.193 &0.393 & 0.422 &0.427&\textcolor{teal}{\bf 0.431} & 0.430 &\textcolor{teal}{\bf 0.431} \\
\cline{2-9}

\hline


\end{tabular}}

\end{center}

\caption{Quantitative ablation studies on the effects of Continual Learning and Rehearsal Rate. We train 7 models with different Rehearsal Rates $R$ and validate the CSD Score and CPC Score on the validation set of each stage. When $R=0$, no Curriculum Learning is applied.  }
\label{table_ablation_rehearsal_rate}

\end{table*}

\subsection{Estimating Style Complexity with FPLID}
We further analyze this behavior using FPLID, the Fokker–Planck Local Intrinsic Dimensionality estimator~\cite{kamkari2024geometric}. For a given disjoint union of manifolds and a point $x$ on this union, the Local Intrinsic Dimensionality (LID) of $x$
 is defined as the dimension of the submanifold that contains $x$, which intuitively reflects the minimal number of variables needed to distinguish $x$ from nearby samples. Higher LID indicates higher image complexity. The FPLID formulation leverages the Fokker–Planck equation to dramatically reduce the computational cost of normal bundle–based estimators~\cite{tempczyk2022lidl,stanczuk2022your}, requiring only a single sample to estimate LID.
 
Concretely, we train a small DDPM U-Net~\cite{ddpm} on target images from $D_{pure}$ and $D_{synth}$ in FLUX VAE\cite{DBLP:journals/corr/KingmaW13} latent space, enabling diffusion-model–based LID estimation for FLUX latents. 
With Variance-preserving DMs, score-matching formation\cite{song2020score} of  FPLID \cite{kamkari2024geometric} is
\begin{equation}
\resizebox{1\linewidth}{!}{%
    \ensuremath{
        \begin{split}
        FPLID(x, t_0) &= D + \big(1 - e^{-B(t_0)}\big) \left( \operatorname{tr} \Big(\nabla s\big(e^{-\tfrac{1}{2}B(t_0)}x, t_0\big)\Big) + \big\Vert s\big(e^{-\tfrac{1}{2}B(t_0)}x, t_0\big) \big\Vert_2^2 \right) \\
        &=  D +\sigma^2(t_0)  \left( \operatorname{tr} \Big(\nabla s\big(\psi(t_0)x, t_0\big)\Big) + \big\Vert s\big(\psi(t_0)x, t_0\big) \big\Vert_2^2 \right) 
        \end{split}
    }%
}
\end{equation}
where $\beta$ is a positive scalar function \citep{song2020score},
\begin{equation}
    \psi(t) = e^{-\tfrac{1}{2} B(t)}, \quad \text{and}\quad \sigma^2(t) = 1 - e^{-B(t)}, 
\end{equation}

\begin{equation}
\quad B(t) \coloneqq \int_0^t \beta(u) d u.
\end{equation}

 Since DDPM \cite{ddpm} and Score Matching \cite{song2020score} could be mutually converted by, ${t}/{T} = t \in [0, 1] \rightarrow {t} \in \{0, 1, \dots, {T}\}$,    $x_{{t}/{T}} \rightarrow x_t $,    $\beta({t}/{T}) = \beta(t) \rightarrow {\beta_t}$, $\psi({t}/{T}) = \psi(t) \rightarrow \sqrt{{\bar{\alpha}_t}}$,  $\sigma({t}/{T}) = \sigma(t) \rightarrow \sqrt{1 - {\bar{\alpha}_t}}$, $ \hat{s}(x, {t}/{T}) = \hat{s}(x, t) \rightarrow -{\epsilon}(x, {t}) /  \sqrt{1 - {\bar{\alpha}_t}}$,\\
where  ${\alpha_t} \coloneqq 1 - {\beta_t}$ and ${\bar{\alpha}_t} \coloneqq \prod_{s=1}^{{t}} {\alpha_t}$.
Thus we could have 
\begin{equation}
{%
    \ensuremath{
        \begin{split}
        FPLID(x, t_0) &  = D - \sqrt{1 - {\bar{\alpha}_{t_0}}} tr \left(\nabla {\epsilon}(\sqrt{{\bar{\alpha}_{t_0}}} x, {t_0})\right) \\
        + \Vert {\epsilon}(\sqrt{{\bar{\alpha}_{t_0}}} x, {t_0})\Vert_2^2,
        \end{split}
    }%
}
\label{eq:method_for_ddpm_optimized}
\end{equation}

where $x$ denotes FLUX VAE Latent of a generated image  from  our SC-DiT,  from $D_{pure}$ and $D_{synth}$, $t_0$ is the timestep for evaluating LID, with $D=16 \times 64 \times 64=65536$. With $\beta_t$ being the Diffusion process hyper-parameter, ${\alpha_t} := 1 - {\beta_t}$, ${\bar{\alpha}_t} := \prod_{s=1}^{{t}} {\alpha_t}$. The notation $tr$ denotes trace operation, $\nabla_x$ denotes  the differentiation operator with respect to $x$ and  $\epsilon$ is the Diffusion UNet. 
\subsection{Relative LID Ranking of Style Clusters }

We demonstrate the qualitative and quantitative ranking of style complexity with our DDPM LID Estimator trained in FLUX VAE Latent Space in the main paper.

To validate the correlation of LID score and human perception of style complexity, we asked 35 users to rank the complexity order of 20 style clusters. The Spearman’s rank-order correlation analysis reveals a  very strong positive monotonic relationship between the FPLID ranking  scores and human judgements with  $\rho = 0.9718$, $p = 0.0007$. This result confirms that the LID ranking order is strongly and significantly aligned with human preferences on style complexity, demonstrating its effectiveness for serving as an indicator for Curriculumn Learning.

\begin{figure*}[!htp]
  \centering
  \hspace{-10mm}
   \includegraphics[width=0.77\linewidth]{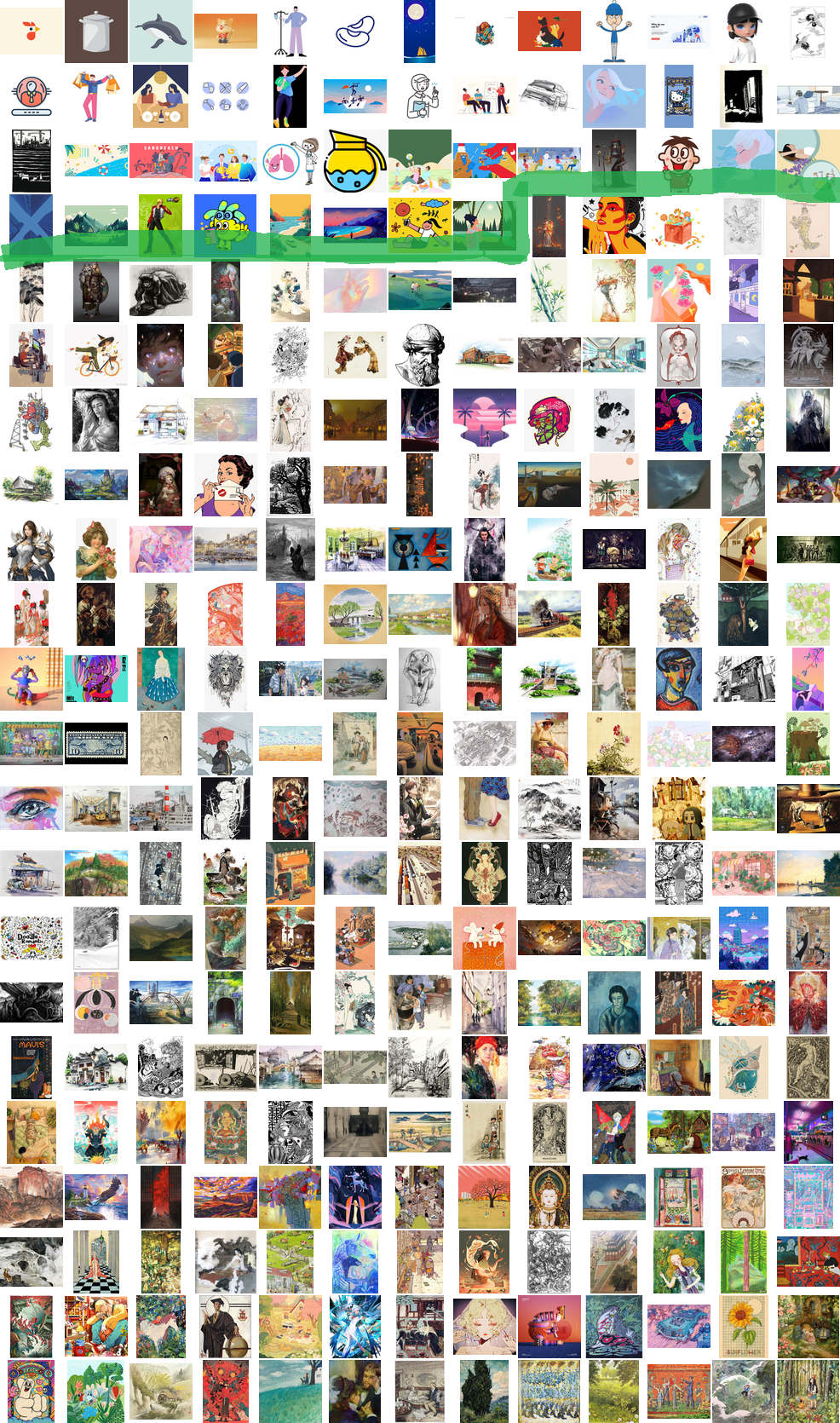}
   \caption{ We randomly select one image each from  hundreds of styles clusters and rank them with Our DDPM LID Estimator in FLUX VAE Latent Space. The LID scores increase from left to right, up to down. }
   \label{figure_lidranking}
\end{figure*}

We show a qualitative example of ranking a subset of the training set. Different with the ranking in the main paper where we control the variance by using the same content and different styles, such strict scheme is impossible for ranking training data. However, shown in Figure \ref{figure_lidranking}, we could still observe the consistent trend from simple to complex in the ranking result, though there are very few outliers. Our manual cut of semantic styles and texture styles locates near the end of the fourth row. 


%


\subsection{Training Triplet Dataset Construction}
\subsubsection{Purified Triplet Matching}
We extract style categories from OmniConsistency dataset \cite{song2025omniconsistency}, which contains 22 style categories from GPT-4O \cite{gpt4o}. We further collect 8 styles with Loras and internet data. We construct the [style ref, content ref, target] triplet with these data and filter the style similarity with CSD score \cite{csd}, content similarity with clip score \cite{Radford2021LearningTV}, facial similarity with arcface \cite{deng2019arcface}.  Previous state-of-the-art style transfer models often suffer from the human facial identity leakage from style reference to content reference. To address such a problem, we purposely set a high proportion of style references to those images containing faces. For simplicity, we call this purified triplet dataset $D_{pure}$ in the following sections. In later experiments, we found that the facial identity leakage problem is alleviated when we train our model with such triplets. Finally we get around 330k training data for $D_{pure}$.  
\subsubsection{Reverse Triplet Synthetic Framework}
Although the purified triplets are clean and of high quality, the scale of such data is small and the data collecting is expensive. In order to utilize in-the-wild style images on internet, we introduce a  reverse triplet synthetic framework inspired by \cite{wang2023stylediffusion}, where we reversely generate a style reference and a content reference from target image. Shown in Figure \ref{figure_triplet}, we use LLM \cite{bai2025qwen2} to generate a prompt bank, where the prompts focus on subjects instead of person because we utilize an internal SDXL \cite{podell2023sdxl} style adapter to generate style reference, which still suffers from facial identity leakage problem aforementioned. We randomly select a prompt from the prompt bank and feed it to the SDXL text encoder.

\paragraph{SDXL-based Style Transfer model to create style reference} This SDXL-based style transfer model CDST \cite{zhang2025cdst} is trained with 14 million text-image pairs, the same data format like common text-to-image models. This model is good at prompt + style reference customization, but does not perform well on content-preserving style transfer with content reference and style reference. 
We feed the style target image to Dinov2 \cite{oquab2023dinov2} to extract image embeddings and input the image embeddings to a style transformer \cite{wang2023styleadapter} to compress the image embeddings to a fixed length tokens. These tokens are then fed to MLP to align the channel dimension and merged into UNet with learnable cross attentions. During inference, we only feed the compressed tokens to the decoder of UNet, which could effectively isolate style features according to Forgedit \cite{zhang2023forgedit,fastimagic}.  Finally, we could get the synthetic style reference.

\paragraph{FLUX-based Photorealistic Converter to create content reference} To get the content reference, we train a specific in-context DiT model shown in Figure \ref{figure_triplet}, which accepts a prompt "make this image photographic" and  a stylized image. It converts this stylized image into a photographic image while preserving the layout and content of the style image. We train this model with  data from $D_{pure}$, where we reform each triplet [style ref, content ref, target] in $D_{pure}$ to [target, content ref] with target being reference image, content reference being target. We introduce an auxiliary Lora in the FLUX structure to learn such editing capability with content RoPE \cite{rope}. The structure and causal attention mechanism is almost the same as SC-DiT in Figure \ref{main}, except that there is just one reference image (the stylized target) instead of two. 

With such synthetic framework, we utilize images in  Style30k\cite{li2024styletokenizer} as style target and reversely synthesize 2 million triplet by generating 20 style reference images and 1 content reference image for each style target image. We further utilize CSD score and CLIP score to filter style similarity and content consistency. Finally we have around 1 million synthetic triplets, denoted $D_{synth}$.
\subsection{Computation Cost}
 Style-CCL-FLUX 1.0 is trained based on FLUX-dev 1.0 \cite{flux2024}, and trained with 4 H100 GPU with 80GB. We trained the model for 200 hours. The inference speed is around 3.6 seconds for generating a $512 \times 512$ image on one H100.

Style-CCL-QIE (QwenStyle and TeleStyle) are also trained with 4 H100 for lora models, 8 H100 for complete parameter SFT.  The resolution is $1024 \times$. With DMD, Style-CCL-QIE genrates an $1024 \times$ image in 4 seconds.



\bibliography{example_paper}
\bibliographystyle{icml2026}


\end{document}